\documentclass[letterpaper, 10 pt, conference]{ieeeconf}  % Comment this line out if you need a4paper

\IEEEoverridecommandlockouts                              % This command is only needed if 
\usepackage{graphicx} % for pdf, bitmapped graphics files
\graphicspath{ {images/} }
\usepackage{amsmath} % assumes amsmath package installed
\usepackage{amsfonts}
\usepackage{amssymb}  % assumes amsmath package installed
\usepackage[utf8]{inputenc} % extended ASCII
\usepackage{placeins}
\usepackage{bm}
\usepackage{blindtext}
\usepackage{cleveref}
\usepackage{lipsum,graphicx,subcaption}
\usepackage{textcomp}
\usepackage{multirow}
\usepackage{xcolor}
\usepackage{algorithm}
\usepackage{algorithmic}

\usepackage{tabularx}

\usepackage[bottom=0.8in, top=0.75in, left=0.75in, right=0.75in]{geometry}

    \newcolumntype{L}{>{\raggedright\arraybackslash}X}

\title{\LARGE \bf
 Dynamic Object Tracking and Masking for Visual SLAM
}

\author{Jonathan Vincent, Mathieu Labb\'{e}, Jean-Samuel Lauzon, Fran\c{c}ois Grondin, \\ Pier-Marc Comtois-Rivet, Fran\c{c}ois Michaud 
\thanks{This work was supported by the Institut du v\'{e}hicule innovant (IVI), Mitacs, Innov\'{E}\'{E} and NSERC. J. Vincent, M. Labb\'{e}, J.-S. Lauzon, F. Grondin and F. Michaud are with the Interdisciplinary Institute for Technological Innovation (3IT), Dept. Elec. Eng. and Comp. Eng., Universit\'{e} de Sherbrooke, 3000 boul. de l'Universit\'{e}, Qu\'{e}bec (Canada) J1K 0A5. P.-M. Comtois-Rivet is with the Institut du V\'{e}hicule Innovant (IVI), 25, boul. Maisonneuve, Saint-J\'{e}rôme, Qu\'{e}bec (Canada), J5L 0A1. {\scriptsize\texttt{\{Jonathan.Vincent2, Mathieu.m.Labbe, Jean-Samuel.Lauzon, Francois.Grondin2, Francois.Michaud\}@USherbrooke.ca, Pmcrivet@ivisolutions.ca}}}%
}

\begin{document}

\maketitle
\thispagestyle{empty}
\pagestyle{empty}

% Abstract
%%%%%%%%%%%%%%%%%%%%%%%%%%%%%%%%%%%%%%%%%%%%%%%%%%%%%%%%%%%%%%%%%%%%%%%%%%%%%%%%
\begin{abstract}

In dynamic environments, performance of visual SLAM techniques can be impaired by visual features taken from moving objects. One solution is to identify those objects so that their visual features can be removed for localization and mapping. This paper presents a simple and fast pipeline that uses deep neural networks, extended Kalman filters and visual SLAM to improve both localization and mapping in dynamic environments (around 14 fps on a GTX 1080). Results on the dynamic sequences from the TUM dataset using RTAB-Map as visual SLAM suggest that the approach achieves similar localization performance compared to other state-of-the-art methods, while also providing the position of the tracked dynamic objects, a 3D map free of those dynamic objects, better loop closure detection with the whole pipeline able to run on a robot moving at moderate speed.

\end{abstract}

% Introduction
%%%%%%%%%%%%%%%%%%%%%%%%%%%%%%%%%%%%%%%%%%%%%%%%%%%%%%%%%%%%%%%%%%%%%%%%%%%%%%%%
\section{INTRODUCTION}
\label{sec:system}

To perform tasks effectively and safely, autonomous mobile robots need accurate and reliable localization from their representation of the environment. Compared to LIDARs (Light Detection And Ranging sensors) and GPS (Global Positioning System), using visual images for Simultaneous Localization and Mapping (SLAM) adds significant information about the environment \cite{fuentes2015visual}, such as color, textures, surface composition that can be used for semantic interpretation of the environment. Standard visual SLAM (vSLAM) techniques perform well in static environments by being able to extract stable visual features from images. However, in environments with dynamic objects (e.g., people, cars, animals), performance decreases significantly because visual features may come from those objects, making localization less reliable \cite{fuentes2015visual}. 
Deep learning architectures have recently demonstrated interesting capabilities to achieve semantic segmentation from images, outperforming traditional techniques in tasks such as image classification \cite{ciregan2012multi}. For instance, Segnet \cite{badrinarayanan2017segnet} is commonly used for semantic segmentation \cite{garcia2017review}. It uses an encoder and a decoder to achieve pixel wise semantic segmentation of a scene. 
%Simultaneous Detection and Segmentation (SDS) \cite{hariharan2014simultaneous} is another technique based on a Multi-scale Combinatorial Grouping (MCG) \cite{arbelaez2014multiscale} process that segments images using the  R-CNN \cite{girshick2014rich} and a Support Vector Machine (SVM). % that generates regions of interest. 
%Features are then extracted from these regions using R-CNN \cite{girshick2014rich}. 
%A trained Support Vector Machine (SVM) classifies those features according to predefined categories, and a non-maximum suppression algorithm refines region segmentation.  
%Other types of architectures, known as Instance Segmentation Networks such as Mask R-CNN \cite{he2017mask}, focus on segmenting independent objects rather than segmenting the whole scene. A region of interest is defined for each objects in the scene and then segmented semantically. 

%Therefore, to identify potential moving objects and to track and mask them during SLAM, this paper presents a novel approach that combines the Mask R-CNN \cite{he2017mask} with Extended Kalman filtering (EKF) in a vSLAM pipeline. 
This paper introduces a simple and fast pipeline that uses neural networks, extended Kalman filters and vSLAM algorithm to deal with dynamic objects. Experiments conducted on the TUM dataset demonstrate the robustness of the proposed method. Our research hypothesis is that a deep learning algorithm can be used to semantically segment object instances in images using \textit{a priori} semantic knowledge of dynamic objects, enabling the identification, tracking and removal of dynamic objects from the scenes using extended Kalman filters to improve both localization and mapping in vSLAM. 
%Mask R-CNN \cite{he2017mask} was developed for instance segmentation. 
%Our approach uses the segmentation output to create a dynamic object mask based on prior knowledge and on the tracking algorithm. This mask is then used to filter out visual features at the input of a vSLAM algorithm. 
By doing so, the approach, referred to as Dynamic Object Tracking and Masking for vSLAM (DOTMask)\footnote{https://github.com/introlab/dotmask} aims at providing six benefits: 1) increased visual odometry performance; 2) increased quality of loop closure detection; 3) produce 3D maps free of dynamic objects; 4) tracking of dynamic objects; 5) modular and fast pipeline.

The paper is organized as follows. Section \ref{sec:relatedwork} presents related work of approaches taking into consideration dynamic objects during localization and during mapping. Section \ref{sec:SSA} describes our approach applied as a pre-processing module to RTAB-Map \cite{labbe2014online}, a vSLAM approach.  Section \ref{sec:expresults} presents the experimental setup, and Section \ref{sec:results} provides comparative results on dynamic sequences taken from the TUM dataset. 

\section{Related Work}
\label{sec:relatedwork}

Some approaches take into consideration dynamic objects during localization. 
For instance, BaMVO \cite{kim2016effective} uses a RGB-D camera to estimate ego-motion. It uses a background model estimator combined with an energy-based dense visual odometry technique to estimate the motion of the camera. Li \textit{et al.} \cite{li2017rgb} developed a static point weighting method which calculates a weight for each edge point in a keyframe. This weight indicates the likelihood of that specific edge point being part of the static environment. Weights are determined by the movement of a depth edge point between two frames and are added to an Intensity Assisted Iterative Closest Point (IA-ICP) method used to perform the registration task in SLAM. 
Sun \textit{et al.} \cite{sun2017improving} present a motion removal approach to increase the localization reliability in dynamic environments. It consists of three steps: 1) detecting moving objects' motion based on ego-motion compensated using image differencing; 2) using a particle filter for tracking; and 3) applying a Maximum-A-Posterior (MAP) estimator on depth images to determine the foreground. This approach is used as the frontend of Dense Visual Odometry (DVO) SLAM \cite{kerl2013dense}. Sun \textit{et al.} \cite{sun2018motion} uses a similar foreground technique but instead of using a MAP they use a foreground model which is updated on-line. All of these approaches demonstrate good localization results using the Technical University of Munich (TUM) dataset \cite{sturm12iros}, however, mapping is yet to be addressed.

SLAM++ \cite{salas2013slam++} and Semantic Fusion \cite{mccormac2017semanticfusion} focus on the mapping aspect of SLAM in dynamic environments. 
SLAM++ \cite{salas2013slam++} is an object-oriented SLAM which achieves efficient semantic scene description using 3D object recognition. %It starts from the premise that the environment has a certain symmetry (or recurrence) in the objects for their identification. 
SLAM++ defines objects using areas of interest to subsequently locate and map them. %SLAM++ demonstrates to improve accurate and dense geometry description over traditional SLAM in an indoor environment. 
However, it needs predefined 3D object models to work. 
Semantic Fusion \cite{mccormac2017semanticfusion} creates a semantic segmented 3D map in real time using RGB-CNN \cite{noh2015learning}, a convolutional deep learning neural network, and a dense SLAM algorithm.
However, SLAM++ and Semantic Fusion do not address SLAM localization accuracy in dynamic environments, neither do they remove dynamic objects in the 3D map.

Other approaches use deep learning algorithm to provide improved localisation and mapping. Fusion++ \cite{mccormac2018fusion++} and MID-Fusion \cite{xu2019mid} uses object-level octree-based volumetric representation to estimate both the camera pose and the object positions. They use deep learning techniques to segment object instances. DynaSLAM \cite{bescos2018dynaslam} proposes to combine multi-view geometry models and deep-learning-based algorithms to detect dynamic objects and to remove them from the images prior to a vSLAM algorithm. They also uses inpainting to recreate the image without object occlusion. DynaSLAM achieves impressive results on the TUM dataset. However, these approaches are not optimized for real-time operation.
  
\section{Dynamic Object Tracking and Masking for vSLAM}
\label{sec:SSA}

\begin{figure}
\centering
\includegraphics[width=0.95\linewidth]{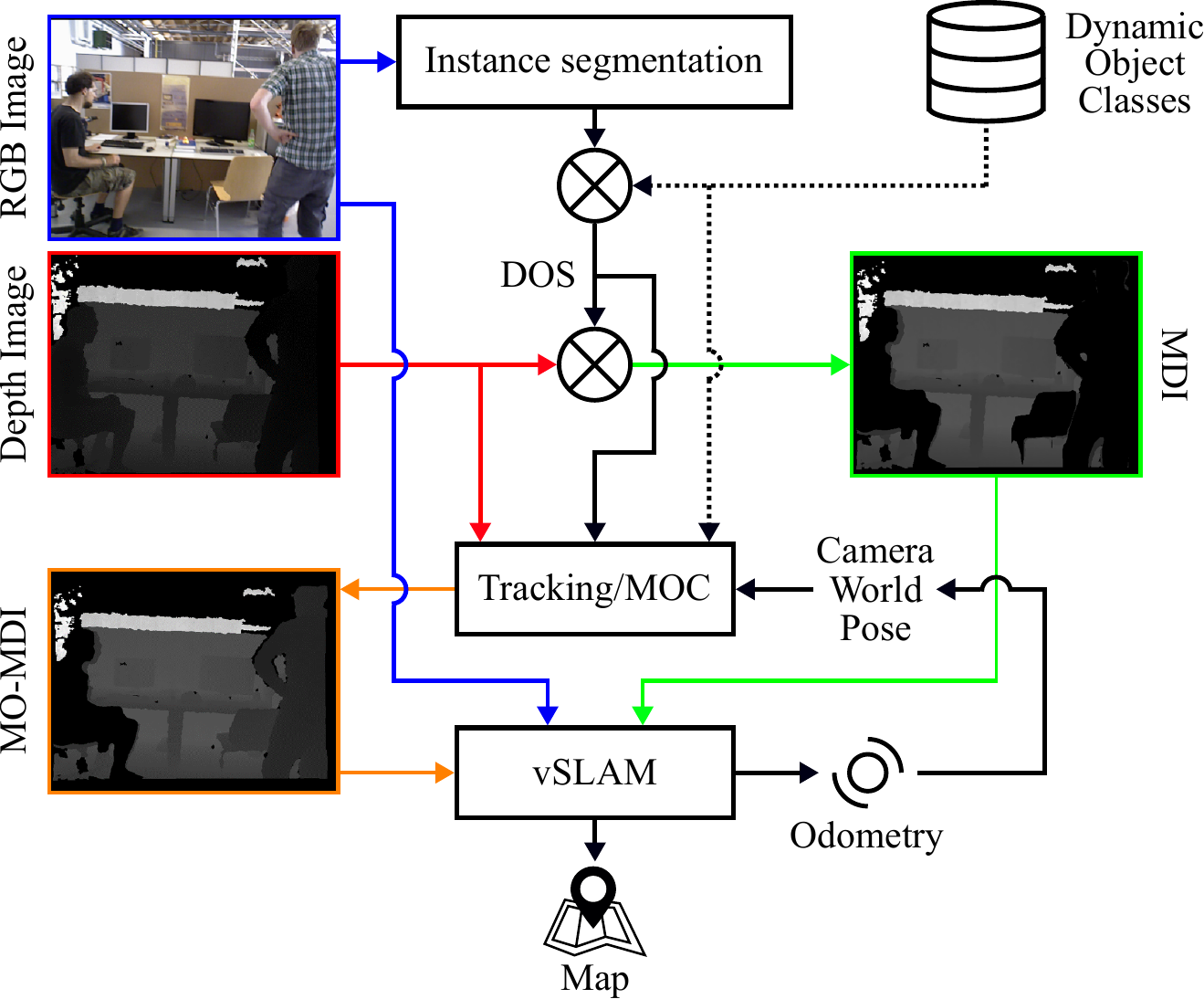}
\caption{Architecture of DOTMask}
\label{gloabal_architecture}
\vspace{-15pt}
\end{figure}

The objective of our work is to provide a fast and complete solution for visual SLAM in dynamic environments. Figure \ref{gloabal_architecture} illustrates the DOTMask pipeline. As a general overview of the approach, a set of objects of interest (OOI) are defined using \textit{a priori} knowledge and understanding of dynamic objects classes that can be found in the environment. Instance segmentation is done using a neural network trained to identify the object classes from an RGB image. %A semantic pixel-wise binary mask is derived from the OOI bounding box and its dynamic object class. 
For each dynamic object instance, its bounding box, class type and binary mask are grouped for convenience and referred as the dynamic object state (DOS). The binary mask of the DOS is then applied to the original depth image, resulting in a masked depth image (MDI). 
The DOS is also sent to the Tracking module. 
After computing a 3D centroid for each masked object, the Tracking module predict the position and velocity of the objects. This information is then used by the Moving Object Classification module (MOC) to classify the object as idle or not based on its class, its estimated velocity and its shape deformation. Moving objects are removed from the original depth image, resulting in the Moving Object Masked Depth Image (MO-MDI). The original RGB image, the MDI and the MO-MDI are used by the vSLAM algorithm. It uses the depth images as a mask for feature extraction thus ignoring features from the masked regions. The MO-MDI is used by the visual odometry algorithm of the vSLAM approach while the MDI is used by both its mapping and loop closure algorithms, resulting in a map free of dynamic objects while still being able to use the features of the idle objects for visual odometry. The updated camera pose is then used in the Tracking module to estimate the position and velocity of the dynamic objects resulting in a closed loop.  

\subsection{Instance Segmentation}

Deep learning algorithms such as Mask R-CNN recently proved to be useful to accomplish instance semantic segmentation \cite{garcia2017review}. A recent and interesting architecture for fast instance segmentation is the YOLACT \cite{yolact-iccv2019} and its update YOLACT++ \cite{yolact-plus-arxiv2019}. This network aims at providing similar results as the Mask-RCNN or the Fully Convolutional Instance-aware Semantic Segmentation (FCIS) \cite{li2017fully} but at a much lower computational cost. YOLACT and YOLACT++ can achieve real-time instance segmentation. Development in neural networks has been incredibly fast in the past few years and probably will be in the years to come. DOTMask was designed the be modular and can easily change the neural network used in the pipeline. In its current state, DOTMask works with Mask-RCNN, YOLACT and YOLACT++. The YOLACT is much faster than the two others and the loss in precision doesn't impact our results. This is why this architecture is used in our tests. The instance segmentation module takes the input RGB image and outputs the bounding box, class and binary mask for each instance. 

%Shown in Fig. \ref{maskrcnn}, 
\iffalse
The Mask R-CNN architecture extends the Faster R-CNN architecture \cite{ren2015faster} by adding a branch for predicting an object mask for semantic segmentation in parallel with the branch for bounding box recognition.
Mask R-CNN is composed of multiple sub-networks: the Convolutional backbone, the Region Proposal Network (RPN), the RoIAlign and the head made of the Box regression and Classification branch, and the Mask branch. 
The Convolutional backbone generates a feature map.
The RPN generates regions of interest in the feature map created by the Convolutional backbone. 
Then, RoIAlign is used to create a fixed size feature map to preserve the exact spatial locations of the regions of interest. 
The Box regression and Classification branch contains a Fully Connected Layer (FCL) which gives box position and classification. 
The Mask branch uses a Fully Convolutional Network (FCN) to create a binary mask for each class instance from the feature map, thus providing instance semantic segmentation.
\fi

\iffalse
\begin{figure*}
\centering
\includegraphics[width=\linewidth]{mask}
\caption{Mask R-CNN architecture} 
\label{maskrcnn}
\end{figure*}
\fi

\subsection{Tracking Using EKF}
Using the DOS from the Instance Segmentation module and odometry from vSLAM, the Tracking module predicts the pose and velocity of the objects in the world frame. This is useful when the camera is moving at speed similar to the objects to track (e.g., moving cars on the highway, robot following a pedestrian) or when idle objects have a high amount of features (e.g., person wearing a plaid shirt).  

First, the Tracking module receives the DOS and the original depth image as a set, defined as $\mathbf{D}^{k} = \{\mathbf{d}^{k}_{1}, ..., \mathbf{d}^{k}_{I}\}$, where $\mathbf{d}^{k}_{i} = \left\{\mathbf{T}^{k}, \mathbf{B}^{k}_{i}, \zeta^{k}_{i}\right\}$ is the object instance detected by the Instance Segmentation module, with $i \in I$, $I = \{1, ..., L\}$, $L$ being the total number of object detection in the frame at time $k$. 
$\mathbf{T} \in \mathbb{R}^{m\times n} $ is the depth image , $\mathbf{B} \in \mathbb{Z}^{m\times n}_2$ is the binary mask and $\mathbf{\zeta} \in J$ is the class ID, with $J = \{1, ..., W\}$, and $W$ is the number of total trained classes in the Instance Segmentation module.

The DOS and the original depth image are used by EKF to estimate the dynamic objects positions and velocities.
EKF provides steady tracking of each object instance corresponding to the object type detected by the neural network. 
An EKF is instantiated for each new object, and \textit{a priori} knowledge from the set of dynamic object classes defines some of the filter's parameters.
This instantiation is made using the following parameters: the class of the object, its binary mask and its 3D centroid position. The 3D centroid is defined as the center of the corresponding bounding box. %, calculated using the function $centroid$.
If the tracked object is observed in the DOS, its position is updated accordingly, otherwise its predicted position using EKF is used. If no observations of the object are made for $e$ number of frames, the object is considered removed from the scene and therefore the filter is discarded. The Tracking module outputs the estimated velocity of the objects to the MOC module. The MOC module will classify the objects as idle or not based on the object class, the filter velocity estimation and the object deformation.%: if the velocity threshold $v$ 
%for a specific object class is exceeded, the object is considered to be moving. 
%The original depth image is then updated to only remove currently moving objects, resulting in the MO-MDI. The MO-MDI is sent to vSLAM odometry to update the camera pose. 

To explain further how the Tracking module works, the following subsections presents in more details the Prediction and Update steps of EKF used by DOTMask.

\subsubsection{Prediction}
Let us define the hidden state $\mathbf{x} \in \mathbb{R}^{6\times 1}$ as the 3D position and velocity of an object referenced in the global map in Cartesian coordinates.
The \textit{a priori} estimate of the state at time $k \in \mathbb{N}$ is predicted based on the previous state at time $k-1$ as in (\ref{eq:state_prediction}):
\begin{equation}
\hat{\mathbf{x}}^{k|k-1} = \mathbf{F} \hat{\mathbf{x}}^{k-1|k-1} \ \textrm{with} \ 
\mathbf{F} = \left[
    \begin{array}{cc}
        \mathbf{I}_{3} & \Delta t\mathbf{I}_{3} \\
        \mathbf{0}_{3} & \mathbf{I}_{3} \\
    \end{array}
    \right]
\label{eq:state_prediction}
\end{equation}
where $\mathbf{F} \in \mathbb{R}^{6\times 6}$ is the state transition matrix, $\Delta t \in \mathbb{R}_+$ is the time between each prediction, $\mathbf{0}_3$ is a $3 \times 3$ zero matrix and $\mathbf{I}_3$ is a $3 \times 3$ identity matrix. Note that the value of $\Delta t$ is redefined before each processing cycle.

The \textit{a priori} estimate of the state covariance ($\mathbf{P}^{k|k-1} \in \mathbb{R}^{6 \times 6}$) at time $k$ is predicted based on the previous state at time $k-1$ as given by (\ref{eq:state_prediction_p}):
\begin{equation}
\mathbf{P}^{k|k-1} = \mathbf{F} \mathbf{P}^{k-1|k-1} \mathbf{F}^T + \mathbf{Q}
\label{eq:state_prediction_p}
\end{equation}
where $\mathbf{Q} \in \mathbb{R}^{6\times 6}$ is the process noise covariance matrix defined using the random acceleration model (\ref{eq:Q}):
\begin{equation}
\mathbf{Q} = \mathbf{\Gamma}\mathbf{\Sigma}\mathbf{\Gamma}^{T} \ \textrm{with} \ \mathbf{\Gamma} = [ 
    \begin{array}{cc}
    \frac{\Delta t^2}{2} \mathbf{I}_{3 \times 3} & \Delta t^2 \mathbf{I}_{3 \times 3}
    \end{array} ]^T
\label{eq:Q}
\end{equation}
where $\mathbf{\Gamma} \in \mathbb{R}^{6\times 3}$ is the mapping between the random acceleration vector $\mathbf{a} \in \mathbb{R}^{3}$ and the state $\mathbf{x}$, and  $\mathbf{\Sigma} \in \mathbb{R}^{3\times 3}$ is the covariance matrix of  $\mathbf{a}$. 
The acceleration components $a_x$, $a_y$ and $a_z$ are assumed to be uncorrelated. 

\iffalse
Thus, the resulting $\mathbf{Q}$ matrix is given by (\ref{eq:innovation}).
%
\begin{equation}
\mathbf{Q} = \left[ 
\arraycolsep=1.4pt\def\arraystretch{1.1}
\begin{array}{cccccc}
\frac{\Delta t^{4}}{4}\sigma^2_{ax} & 0 & 0 & \frac{\Delta t^{4}}{4}\sigma^3_{ax} & 0 & 0 \\
0 & \frac{\Delta t^{4}}{4}\sigma^2_{ay} & 0 & 0 & \frac{\Delta t^{4}}{4}\sigma^3_{ay} & 0 \\
0 & 0 & \frac{\Delta t^{4}}{4}\sigma^2_{az} & 0 & 0 & \frac{\Delta t^{4}}{4}\sigma^3_{az} \\
\frac{\Delta t^{4}}{4}\sigma^3_{ax} & 0 & 0 & \Delta t^{2}\sigma^2_{ax} & 0 & 0 \\
0 & \frac{\Delta t^{4}}{4}\sigma^3_{ay} & 0 & 0 & \Delta t^{2}\sigma^2_{ay} & 0 \\
0 & 0 & \frac{\Delta t^{4}}{4}\sigma^3_{az} & 0 & 0 & \Delta t^{2}\sigma^2_{az} \\
\end{array}
\right]
\label{eq:innovation}
\end{equation}
%
\fi

The dynamic of every detected objects may vary greatly depending on its class. For instance, a car does not have the same dynamic as a mug. To better track different types of objects, a covariance matrix is defined for each class to better represent their respective process noise. 

\subsubsection{Update}

In EKF, the Update step starts by evaluating the innovation $\Tilde{\mathbf{y}}^{k}$ defined as (\ref{eq:innovation_2}): 
\begin{equation}
\Tilde{\mathbf{y}}^{k} = \mathbf{z}^{k} - \mathbf{\hat{h}}^{k}( \hat{\mathbf{x}}^{k|k-1})
\label{eq:innovation_2}
\end{equation}
where $\mathbf{z}^{k} \in \mathbb{R}^3$ is a 3D observation of a masked object in reference to the camera for each object instance, with $\mathbf{z} = [z_x\ z_y\ z_z]^T$, $z_{x} = (\mu_{x}-C_{x})z_{z}/f_{x}$ and $z_{y} = (\mu_{y}-C_{y})z_{z}/f_{y}$, where ${C_x}$ and ${C_y}$ are the principal center point coordinate and ${f_x}$ and ${f_y}$ are the focal lengths expressed in pixels. 
${z_{z}}$ is approximated using the average depth from the masked region on the depth image. The expressions $\mu_x$ and $\mu_y$ stand for the center of the bounding box.

To simplify the following equations, ($s$, $c$) represent respectively the sine and cosine operations of the the Euler angles $\phi$, $\theta$, $\psi$ (roll, pitch, yaw). 
$h(\mathbf{x}^k) \in \mathbb{R}^4$ is the observation function which maps the true state space $\mathbf{x}^k$ to the observed state space $\mathbf{z}^k$. $\mathbf{\hat{h}}(\mathbf{x}^k)$ is the three first terms of $\mathbf{h}(\mathbf{x}^k)$. However, in our case, the transform between those spaces is not linear, justifying the use of EKF. 
\iffalse
$\mathbf{H}^{k}$ is then the following jacobian : 
\begin{equation}
\mathbf{H}^{k} = \left.\frac{\delta{h(\mathbf{x},\mathbf{z}^{k})}}{\delta{x}}\right|_\hat{\mathbf{x}}^{k|k-1}
\label{eq:observation_model_jacobian}
\end{equation}
\fi
The non-linear rotation matrix used to transform the estimate state $\mathbf{\Hat{x}^k}$ in the observed state $\mathbf{z}^k$ follows the ($x,y,z$) Tait-Bryan convention and is given by $h(\mathbf{\Hat{x}}^k) = [h_\phi\ h_\theta\ h_\psi\ 1]$, where: 
\begin{equation}
\small
\hspace{-6pt}
\begin{array}{c}
h_\phi = (c_\phi c_\theta)\Hat{x}_x+(c_\phi s_\theta s_\psi - c_\psi s_\phi)\Hat{x}_y + (s_\phi s_\psi + c_\phi c_\psi s_\theta)\Hat{x}_z+c_x\\
h_\theta =(c_\theta s_\phi)\Hat{x}_x+(c_\phi c_\psi + s_\phi s_\theta s_\psi)\Hat{x}_y+(c_\psi s_\phi s_\theta - c_\phi s_\psi)\Hat{x}_z+c_y\\
h_\psi = -(s_\theta)\Hat{x}_x+(c_\theta s_\psi)\Hat{x}_y+(c_\theta c_\psi)\Hat{x}_z+c_z\\
\end{array}
\label{eq:rotation_h3}
\end{equation}
and $c_x$, $c_y$ and $c_z$ are the coordinate of the camera referenced to the world, which is derived using vSLAM odometry. 

The innovation covariance $\mathbf{S}_{k} \in \mathbb{R}^{3 \times 3}$ is defined as follows, where the expression $\mathbf{H}^{k} \in \mathbb{R}^{3\times 6}$ stands for the Jacobian of $h(\hat{\mathbf{x}}^k)$:

\begin{equation}
\mathbf{S}^{k} = \mathbf{H}^{k} \mathbf{P}^{k|k-1} \left(\mathbf{H}^{k}\right)^T + \mathbf{R}^{k}
\label{eq:innovation_cov}
\end{equation}
where $\mathbf{R}^{k} \in \mathbb{R}^{3\times 3}$ is the covariance of the observation noise, its diagonal terms stand for  the imprecision of the RGB-D camera. 
The near optimal Kalman gain $\mathbf{K}^k \in \mathbb{R}^{3\times 3}$ is defined as follows:
\begin{equation}
\mathbf{K}^{k} = \mathbf{P}^{k|k-1} \left(\mathbf{H}^{k}\right)^T({\mathbf{S}^{k}})^{-1}
\label{eq:kalman_gain}
\end{equation}

Finally, the updated state estimate $\hat{\mathbf{x}}^{k|k}$ and the covariance estimate are given respectively by (\ref{eq:up_state_estimate_1}) and (\ref{eq:up_state_estimate_2}).
\begin{equation}
\hat{\mathbf{x}}^{k|k} = \hat{\mathbf{x}}^{k|k-1} +  \mathbf{K}^{k} \Tilde{\mathbf{y}}^{k}
\label{eq:up_state_estimate_1}
\end{equation}
\begin{equation}
\mathbf{P}^{k|k} = ( \mathbf{I}_6 -  \mathbf{K}^{k} \mathbf{H}^{k} )\mathbf{P}^{k|k-1} 
\label{eq:up_state_estimate_2}
\end{equation}

\subsection{Moving Object Classification}

The MOC module classify dynamic objects as either moving or idle. It takes as inputs the dynamic objects class, velocity and mask. The object velocity comes from the tracking module estimation. The object class and mask are directly obtained from the DOS. The object class defines if the object is rigid or not. The deformation of non-rigid object is computed using the intersection over union (IoU) of the masks of the object at time $k$ and $k-1$. The IoU algorithm takes two arbitrary convex shape $\mathbf{M}^{k-1}, \mathbf{M}^{k}$ and is defined as $\textrm{IoU} = |\mathbf{M}^{k}\cap\mathbf{M}^{k-1}| / |\mathbf{M}^{k}\cup\mathbf{M}^{k-1}|$,  where $|\dots|$ is the cardinality of the set. A dynamic object is classified as moving if its velocity is higher than a predefined threshold or if it is an non-rigid object with an IoU above another predefined threshold. The original depth image is then updated resulting in the MO-MDI. The MO-MDI is sent to the vSLAM odometry to update the camera pose.

\iffalse
\begin{algorithm}
\caption{Moving object classification}
\begin{algorithmic}
\REQUIRE
\ENSURE

\IF{$n < 0$}
\ENDIF
\end{algorithmic}
\label{alg:moving_object_classification}
\end{algorithm}
\fi

\section{EXPERIMENTAL SETUP}
\label{sec:expresults}

To test our DOTMask approach, we chose to use the TUM dataset because it presents challenging indoor dynamic RGB-D sequences with ground truth to evaluate visual odometry techniques. Also, TUM is commonly used to compare with other state-of-the-art techniques. We used sequences in low dynamic and highly dynamic environments.

\iffalse
The sequences used are:

\begin{itemize}
\item \textit{fr2/desk}: the camera moves around a desk.
\item \textit{fr2/long\_household}: the camera makes a circle around a desk.
\item \textit{fr3/sitting\_static}: the camera is held in place while two persons sit in an office scene.
\item \textit{fr3/sitting\_xyz}: the camera moves along the $x$,$y$ and $z$ axis while two persons sit in an office scene.
\item \textit{fr3/sitting\_rpy}: the camera rotates along the $x$,$y$ and $z$ axis while two person sit in an office scene.
\item \textit{fr3/sitting\_halfsphere}: the camera moves on a small half sphere of approximately 1 m of diameter while two persons sit in an office scene.
\item \textit{fr2/desk\_with\_person}: the camera moves around a desk while a person walk by and sit at the desk.
\item \textit{fr3/walking\_static}: the camera is held in place while two persons walk in an office scene.
\item \textit{fr3/walking\_xyz}: the camera moves along the $x$,$y$ and $z$ axis while two persons walk in an office scene.
\item \textit{fr3/walking\_rpy}: the camera rotates along the $x$,$y$ and $z$ axis while two persons walk in an office scene.
\item \textit{fr3/walking\_halfsphere}: the camera moves on a small half sphere of approximately 1 m of diameter while two persons walk in an office scene.
\end{itemize}
\fi

For our experimental setup, ROS %\footnote{http://www.ros.org} 
is used as a middleware to make the interconnections between the input images, segmentation network, EKF and RTAB-Map.
The deep learning library PyTorch %\cite{NEURIPS2019_9015} 
is used for the instance segmentation algorithm. % because of its simplicity, performance and large user base. 
The ResNet-50-FPN backbone is used for the YOLACT architecture because this configuration achieves the best results at a higher framerate \cite{yolact-iccv2019}.  
Our Instance segmentation module is based on the implementation of YOLACT by dbolya\footnote{https://github.com/dbolya/yolact} and its pre-trained weights. The network is trained on all 91 classes of the COCO dataset. The COCO dataset is often used to compare state-of-the-art instance segmentation approaches, which is why we chose to use it in our trials. In our tests, person, chair, cup and bottle are the the OOI used because of their presence in the TUM dataset and in our in-house tests.The RTAB-Map library \cite{labbe2014online} is also used, which includes various state-of-the-art visual odometry algorithms, a loop closure detection approach and a 3D map render.

Table \ref{table:params} presents the parameters used for DOTMask in our trials, based on empirical observations in the evaluated TUM sequences and our understanding of the nature of the objects. 
A probability threshold $p$ and a maximum instance number $m$ are used to reduce the number of object instances to feed into the pipeline. Only detections with a score above $p$ are used and at maximum, $m$ objects detections are processed. This provides faster and more robust tracking.
\begin{table}
\centering
\caption{Experimental Parameters}
\renewcommand*{\arraystretch}{1.1}
\begin{tabular}[width=0.4\linewidth]{ |c||c|}
 \hline
Description & Value \\
  \hline
Frame to terminate object tracking & 10 \\
%Dilatation ($d$) & 20 \\
Score threshold ($s$) & 0.1 \\
Maximum number of observations ($m$) & 5 \\
Velocity threshold for a person & 0.01 m/sec \\
Velocity threshold for the other objects & 0.1 m/sec \\
Random acceleration for a person & 0.62 m/s\textsuperscript{2}\\
Random acceleration for other objects & 1.0 m/s\textsuperscript{2} \\
 \hline
\end{tabular}
\label{table:params}
\vspace{-10pt}
\end{table}

\begin{figure}
  \centering
  \begin{subfigure}{\columnwidth}
  \centering
  \includegraphics[width=0.445\linewidth]{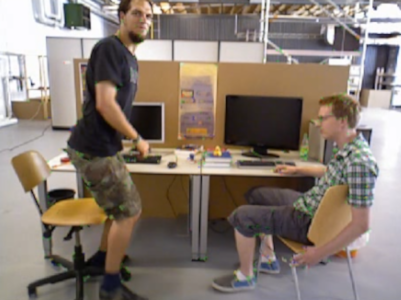}
  \includegraphics[width=0.445\linewidth]{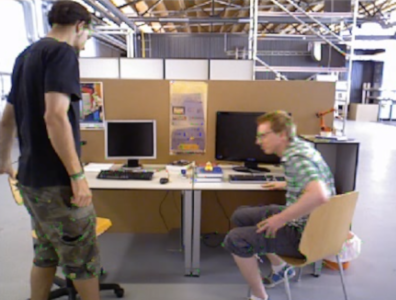}
  \caption{Original RGB Image}
  \label{fig:rtabmap_features_a}
  \vspace{5pt}
  \end{subfigure}
  \begin{subfigure}{\columnwidth}
  \centering
  \includegraphics[width=0.44\linewidth]{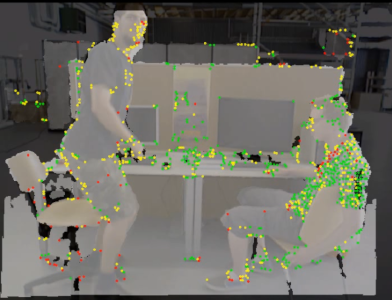}
  \includegraphics[width=0.454\linewidth]{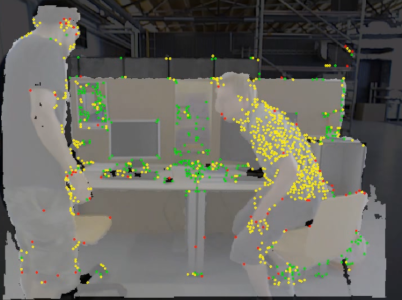}
  \caption{RGB and depth image superposed without DOTMask}
  \label{fig:rtabmap_features_b}
  \vspace{5pt}
  \end{subfigure}
  \begin{subfigure}{\columnwidth}
  \centering
  \includegraphics[width=0.44\linewidth]{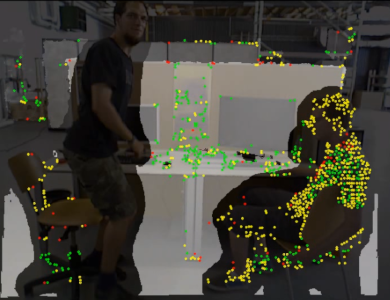}
  \includegraphics[width=0.454\linewidth]{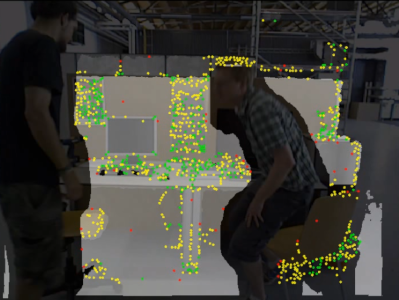}
  \caption{RGB and depth image superposed with DOTMask}
  \label{fig:rtabmap_features_c}
  \end{subfigure}
  \caption{RTAB-Map features (colored dots) not appearing on moving objects with DOTMask}
  \label{fig:rtabmap_features}
  \vspace{-20pt}
\end{figure}

\section{RESULTS}
\label{sec:results}
\begin{figure*}
\begin{tabular}{c c c}
  \centering
  \begin{subfigure}{0.29\textwidth}
  \includegraphics[width=\linewidth]{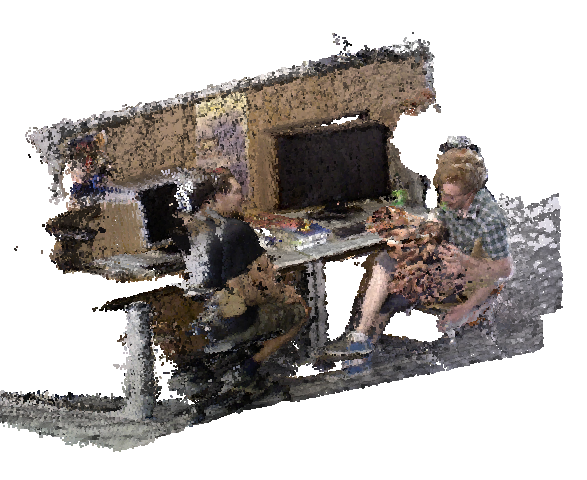}
  \par
  \vspace{0.1cm}
  \rule{0pt}{1ex} \includegraphics[width=\linewidth]{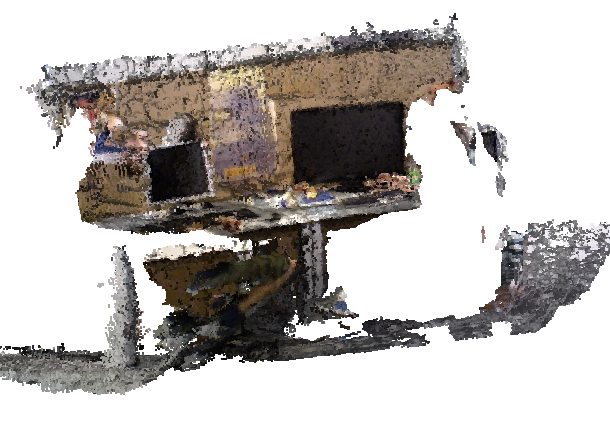}
  \rule{0pt}{3ex} \caption{fr3/sitting\_static}
  \end{subfigure} 
  &
  \begin{subfigure}{0.29\textwidth}
  \includegraphics[width=\linewidth]{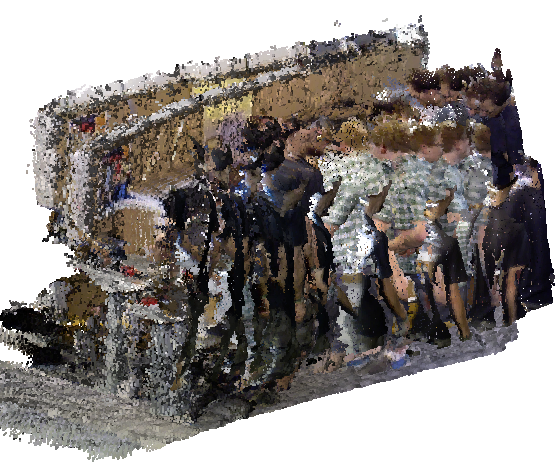}
  \par
  \vspace{0.1cm}
  \includegraphics[width=\linewidth]{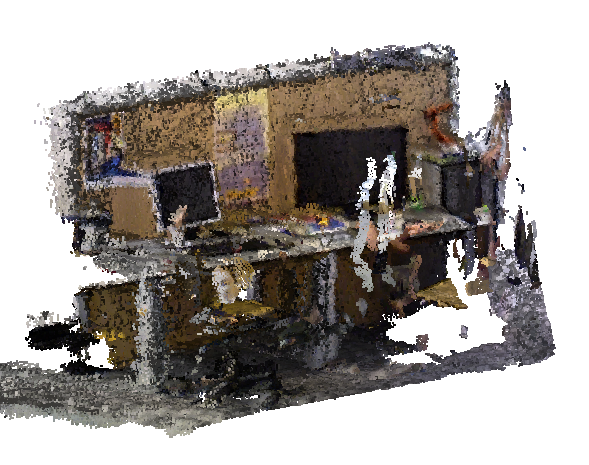}
  \caption{fr3/walking\_static}
  \end{subfigure}
  &
  \begin{subfigure}{0.29\textwidth}
  \includegraphics[width=\linewidth]{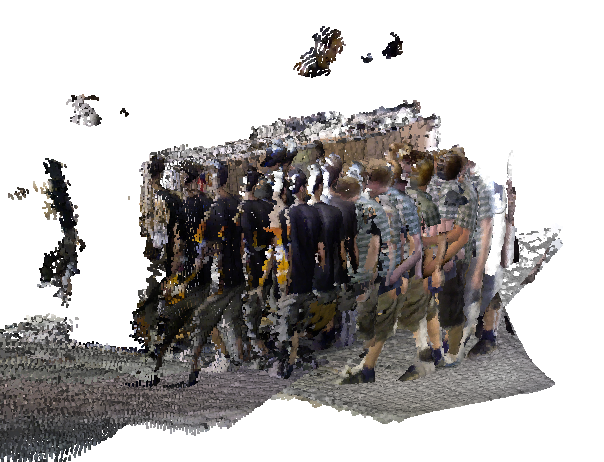}
  \par
  \vspace{0.1cm}
  \includegraphics[width=\linewidth]{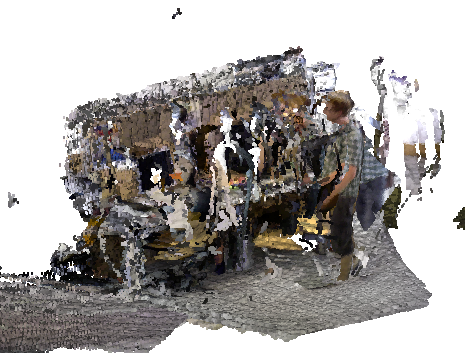}
  \rule{0pt}{3ex}\caption{fr3/walking\_rpy}
  \end{subfigure}
\end{tabular}
\caption{RTAB-Map 3D rendered map from the TUM sequences, without (top) and with (bottom) DOTMask}
\label{3dmap_1}
\vspace{-10pt}
\end{figure*}

Trials were conducted in comparison with approaches by Kim and Kim \cite{kim2016effective}, Sun \textit{et al.} \cite{sun2017improving}, Bescos \textit{et al.} \cite{bescos2018dynaslam} and RTAB-Map, the latter being also used with DOTMask. Figure \ref{fig:rtabmap_features_a} shows two original RGB frames in the TUM dataset, along with their superimposed RGB and depth images with features used by RTAB-Map (Fig. \ref{fig:rtabmap_features_b}) and with DOTMask (Fig. \ref{fig:rtabmap_features_c}). 
Using the depth image as a mask to filter outlying features, dynamic objects (i.e., humans and chairs in this case) are filtered out because the MDI includes the semantic mask. 
The MO-MDI is used by RTAB-Map to compute visual odometry, keeping only the features from static objects as seen in Fig. \ref{fig:rtabmap_features_c} (left vs right) with the colored dots representing visual features used for visual odometry. In the left image of Fig. \ref{fig:rtabmap_features_c}, the man on the left is classified by the Tracking module as moving, while the man on the right is classified as being idle, resulting in keeping his visual features. 
In the rigth image of Fig. \ref{fig:rtabmap_features_c}, the man on the right is also classified as moving because he is standing up, masking his visual features. Figure \ref{3dmap_1} illustrates the influence of MDI, which contains the depth mask of all the dynamic objects, either idle or not, to generate a map free of dynamic objects. This has two benefits: it creates a more visually accurate 3D rendered map, and it improves loop closure detection. 
The differences in the 3D generated maps between RTAB-Map without and with DOTMask are very apparent: there are less artifacts of dynamic objects and less drifting. The \textit{fr3/walking\_static} sequence shows improved quality in the map, while the \textit{fr3/walking\_rpy} sequence presents some undesirable artifacts. These artifacts are caused either by the mask failing to identify dynamic objects that are tilted or upside down or by the time delay between the RGB image and its corresponding depth image. 
The \textit{fr3/sitting\_static} shows the result when masking idle object, resulting in completely removing the dynamic objects from the scene.

Table \ref{table:atermse} characterizes the overall SLAM quality in terms of absolute trajectory error (ATE). 
In almost all cases, DOTMask improves the ATE compared to RTAB-Map alone (as seen in the last column of the table). Table \ref{table:atermse} characterizes the overall SLAM quality in terms of absolute trajectory error (ATE). 
%In almost all cases, DOTMask improves the ATE compared to RTAB-Map alone (as seen in the last column of the table). 
While DynaSLAM is better in almost every sequences, DOTMask is not far off with closer values compared to the other techniques. 
%While DynaSLAM is better in almost every sequences, DOTMask is not far off with closer values compared to the other techniques. 

\begin{table}
\centering
\caption{Absolute Transitional Error (ATE) RMSE in cm}
\renewcommand*{\arraystretch}{1.15}
\begin{tabular}[width=\linewidth]{ |l||c|c|c||c|c||c| }
\cline{1-7}
TUM Seqs & \rotatebox[origin=c]{90}{BaMVO} & \rotatebox[origin=c]{90}{\ Sun et al.\ } & \rotatebox[origin=c]{90}{DynaSLAM} & \rotatebox[origin=c]{90}{RTAB-Map} & \rotatebox[origin=c]{90}{\ DOTMask\ } &  \rotatebox[origin=c]{90}{\ Impr. (\%)\ }\\
 \hline
 \hline
fr3/sit\_static & 2.48 & - & - & 1.70 & \textbf{0.60} & 64.71 \\
fr3/sit\_xyz & 4.82 & 3.17 & \textbf{1.5} & 1.60 & 1.80 & -12.50 \\
 \hline
fr3/wlk\_static & 13.39 & \textbf{0.60} & 2.61 & 10.7 & 0.80 & 92.52 \\
fr3/wlk\_xyz & 23.26 & 9.32 & \textbf{1.50} & 24.50 & 2.10 & 91.42 \\
fr3/wlk\_rpy & 35.84 &13.33 & \textbf{3.50} & 22.80 &  5.30 & 76.75 \\
fr3/wlk\_halfsph & 17.38 & 12.52 & \textbf{2.50} & 14.50 & 4.00 & 72.41 \\
 \hline
\end{tabular}
\label{table:atermse}
\end{table}

Table \ref{table:loopdetect} presents the number of loop closure detections, the mean translation error ($T_{err}$) and the mean rotational error ($R_{err}$) on each sequences both with and without DOTMask. In all sequences, DOTMask helps RTAB-Map to make more loop closures while also lowering both mean errors. Since loop closure features are computed from the depth image (MDI), using DOTMask forces RTAB-Map to use only features from static object hence providing better loop closures. 

On the \textit{fr3/sitting\_xyz} sequence, RTAB-Map alone provides better performance in both ATE and loop closure detection. In this entire sequence, the dynamic objects do not move. While the MO-MDI enables features from idle dynamic objects to be used by the odometry algorithm, the MDI does not enables those same features for the loop closure algorithm. Since nothing is moving in this particular sequence, all features will help to provide a better localisation. However, this case is not representative of dynamic environments.

\begin{table}
\centering
\caption{Loop Closure Analysis}
\renewcommand*{\arraystretch}{1.15}
\begin{tabular}[width=0.4\linewidth]{ |l||c|c|c||c|c|c|}
 \hline
  & \multicolumn{3}{ c|| }{RTAB-Map} & \multicolumn{3}{ c| }{DOTMask }\\
 \hline
  TUM Seqs & Nb & $T_{err}$ & $R_{err}$ & Nb & $T_{err}$ & $R_{err}$ \\
   & loop & (cm) & (deg) & loop & (cm) & (deg) \\
 \hline
 \hline
 fr3/sit\_static & 33 & 1.80 & 0.26 & \textbf{1246} & \textbf{0.60} & \textbf{0.21} \\
 fr3/sit\_xyz & 288 & \textbf{2.10} & \textbf{0.42} & \textbf{1486} & 2.50 & 0.45 \\
 \hline
 fr3/wlk\_static & 105 & 9.00 & 0.18 & \textbf{1260} & \textbf{7.00} & \textbf{0.15}  \\
 fr3/wlk\_xyz & 55 & 6.5 & 0.99 & \textbf{1516} & \textbf{2.9} & \textbf{0.45}  \\
 fr3/wlk\_halfs. & 121 & 5.90 & 0.84 & \textbf{964} & \textbf{4.90} & \textbf{0.79} \\
 fr3/wlk\_rpy & 94 & 6.7 & 1.06 & \textbf{965} & \textbf{6.00} & \textbf{1.04} \\
 \hline
\end{tabular}
\label{table:loopdetect}
\vspace{-10pt}
\end{table}
\normalsize
\begin{table}
\centering
\caption{Timing Analysis}
\renewcommand*{\arraystretch}{1.15}
\begin{tabular}[width=\linewidth]{ |l||c|c|c|c|c| }
\cline{1-5}
 Aproach & Img. Res. & Avg. Time & CPU &  GPU\\
 \hline
 \hline
BaMVO. & 320$\times$240 & 42.6 ms & i7 3.3GHz & - \\
Sun et al. & 640$\times$480 & 500 ms & i5 & - \\
DynaSLAM & 640$\times$480 & 500 ms & - & - \\
DOTMask & 640$\times$480 & 70 ms & i5-8600K & GTX1080 \\
DOTMask & 640$\times$480 & 125 ms & i7-8750H & GTX1050 \\
 \hline
\end{tabular}
\label{table:timing}
\vspace{-10pt}
\end{table}

Table \ref{table:timing} presents the average computation time to process a frame for each approach without vSLAM and odometry algorithms. Results are processed on a computer equipped with a GTX 1080 GPU and a I5-8600k CPU. DOTMask was also tested on a laptop with a GTX 1050 where it achieved an average of 8 frames per second. At 70 ms, it can run on a mobile robot operating at a moderate speed. The fastest method is BaMVO with only 42 ms cycle time.

%\begin{figure}
%  \centering
%  \begin{subfigure}{0.32\columnwidth}
%  \includegraphics[width=\linewidth]{test_js_01.jpg}
%  \caption{$t=3$ sec}
%  \label{fig:test_js_01}
%  \end{subfigure} 
%  \begin{subfigure}{0.32\columnwidth}
%  \includegraphics[width=\linewidth]{test_js_02.jpg}
%  \caption{$t=7$ sec}
%  \label{fig:test_js_02}
%  \end{subfigure}
%  \begin{subfigure}{0.32\columnwidth}
%  \includegraphics[width=\linewidth]{test_js_03.jpg}
%  \caption{$t=10$ sec}
%  \label{fig:test_js_03}
%  \end{subfigure}
%\caption{Frames from the test scenario in a real environment}
%\label{fig:testreal}
%\end{figure}

\begin{figure}
\centering
\includegraphics[width=0.8\linewidth]{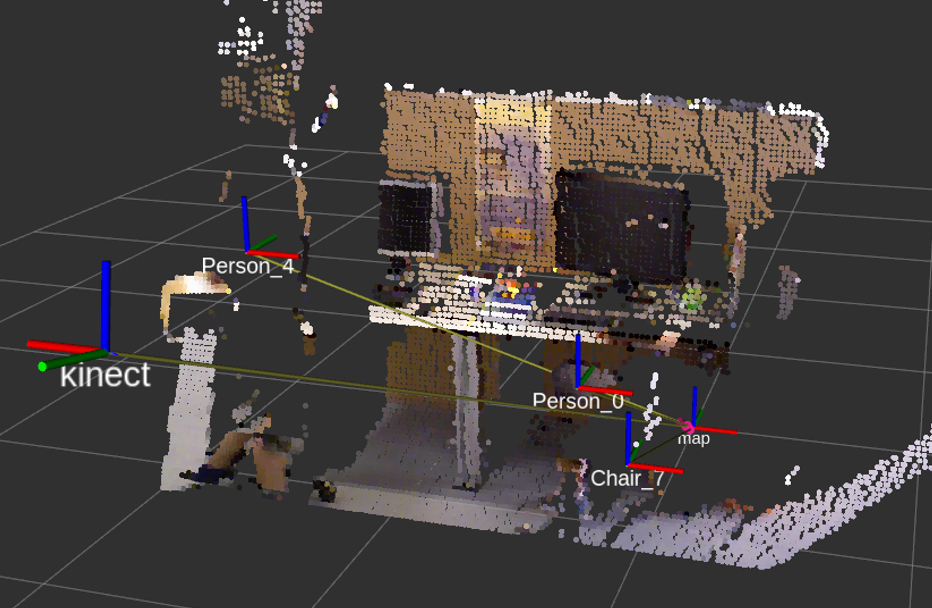}
\caption{Position of tracked dynamic objects shown in RVIZ}
\label{fig:tfros}
\vspace{-5pt}
\end{figure}

\begin{figure}
  \centering
  \begin{subfigure}{0.485\columnwidth}
  \includegraphics[width=\linewidth]{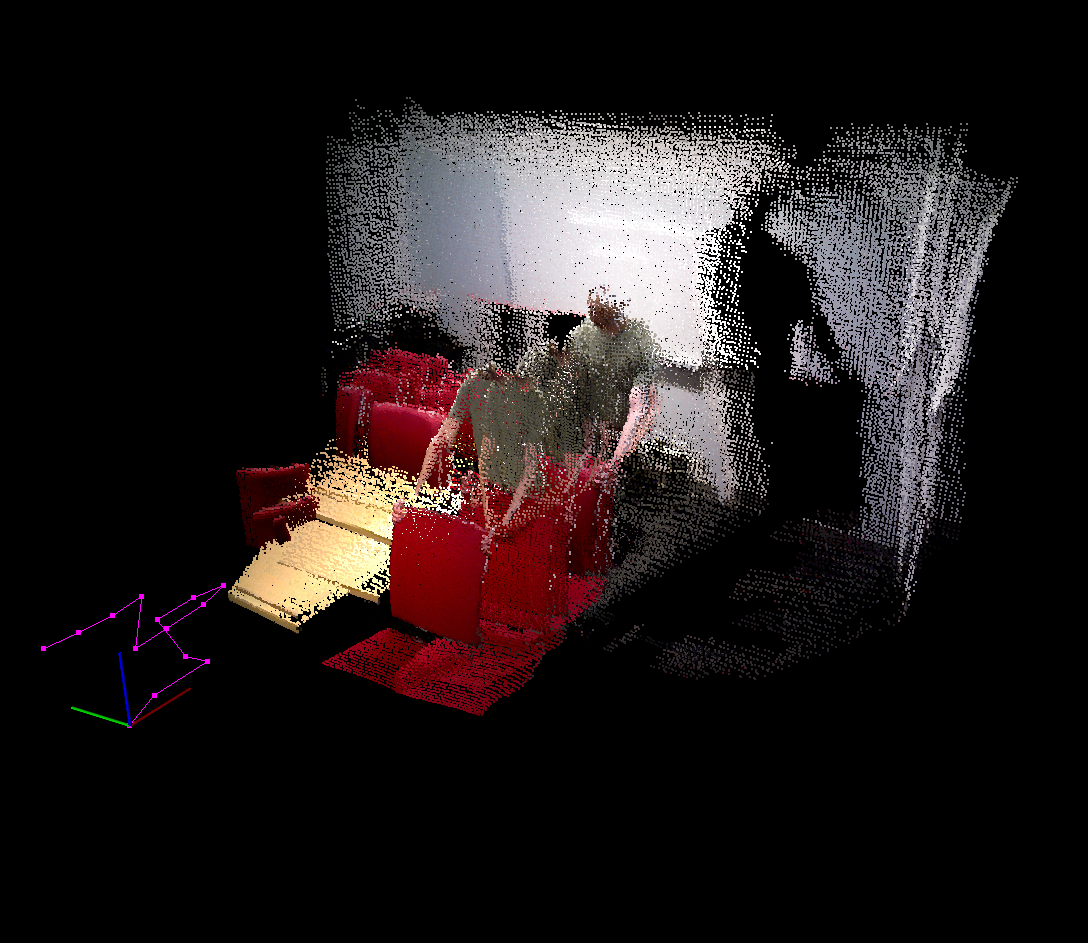}
  \caption{RTAB-Map alone}
  \label{fig:test_js_01}
  \end{subfigure} 
  \begin{subfigure}{0.49\columnwidth}
  \includegraphics[width=\linewidth]{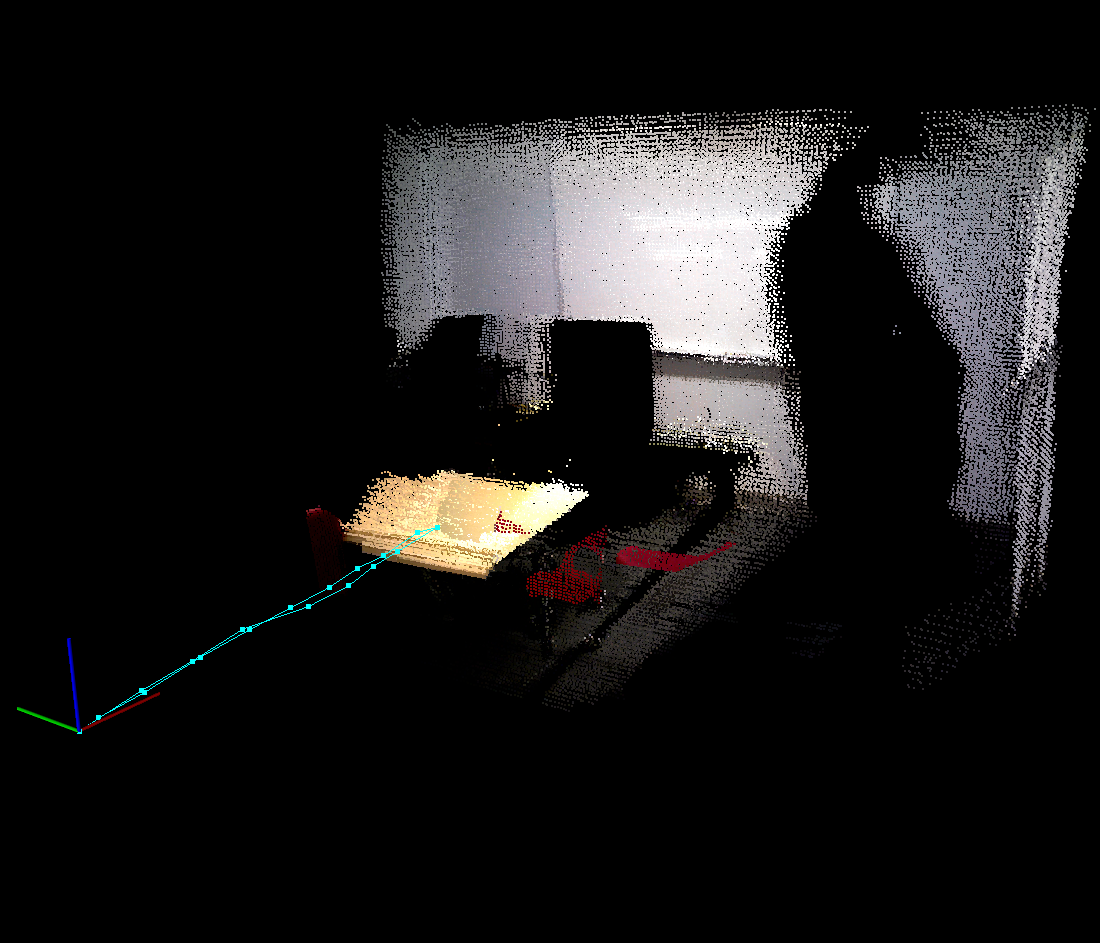}
  \caption{RTAB-Map with DOTMask}
  \label{fig:test_js_02}
  \end{subfigure}
\caption{3D map and odometry improved with DOTMask}
\label{fig:realresults}
\vspace{-15pt}
\end{figure}

Figure \ref{fig:tfros} shows the tracked dynamic objects in the ROS visualizer RViz. DOTMask generates ROS transforms to track the position of the objects. Those transforms could easily be used in other ROS applications. Figure \ref{fig:realresults} shows the difference between RTAB-Map and DOTMask in a real scene where a robot moves at a similar speed as dynamic objects (chairs and humans). The pink and blue lines represent the odometry of RTAB-Map without and with DOTMask. These results suggest qualitatively that DOTMask improves the odometry and the 3D map. 

\section{CONCLUSION}
\label{sec:conclusion}    

This paper presents DOTMask, a fast and modular pipeline that uses a deep learning algorithm to semantically segment images, enabling the tracking and masking of dynamic objects in scenes to improve both localization and mapping in vSLAM. Our approach aims at providing a simple and complete pipeline to allow mobile robots to operate in dynamic environments. Results on the TUM dataset suggest that using DOTMask with RTAB-Map provides similar performance compared to other state-of-the-art localization approaches while providing an improved 3D map, dynamic objects tracking and higher loop closure detection. While DOTMask does not outperform DynaSLAM on the TUM dataset or outrun BaMVO, it reveals to be a good compromise for robotic applications. Because DOTMask pipeline is highly modular, it can also evolve with future improvements of deep learning architectures and new sets of dynamic object classes. In future work, we want to use the tracked dynamic objects to create a global 3D map with object permanence, and explore more complex neural networks\footnote{https://github.com/daijucug/Mask-RCNN-TF\_detection-human\_segment-body\_keypoint-regression} to add body keypoint tracking, which could significantly improve human feature extraction. We would also like to explore techniques to detect outlier segmentations from the neural network to improve robustness.

% Bibliography
%%%%%%%%%%%%%%%%%%%%%%%%%%%%%%%%%%%%%%%%%%%%%%%%%%%%%%%%%%%%%%%%%%%%%%%%%%%%%%%%
\bibliographystyle{IEEEtran}
\bibliography{IEEEabrv,references}

\end{document}